\documentclass[lettersize,journal]{IEEEtran}
\usepackage{amsmath,amsfonts}
\usepackage{algorithmic}
\usepackage{array}
\usepackage[caption=false,font=normalsize,labelfont=sf,textfont=sf]{subfig}
\usepackage{textcomp}
\usepackage{stfloats}
\usepackage{url}
\usepackage{verbatim}
\usepackage{graphicx}
\usepackage[table]{xcolor}
\usepackage{multirow}
\usepackage{algorithm}
\def\BibTeX{{\rm B\kern-.05em{\sc i\kern-.025em b}\kern-.08em
    T\kern-.1667em\lower.7ex\hbox{E}\kern-.125emX}}
\usepackage{balance}
\usepackage[colorlinks=true,citecolor=green]{hyperref}

\def\etal{\emph{et al.} }

\begin{document}
\title{Dual Teacher Knowledge Distillation with Domain Alignment for Face Anti-spoofing}


\author{
Zhe Kong, Wentian Zhang, Tao Wang, Kaihao Zhang, Yuexiang Li, Xiaoying Tang, Wenhan Luo
\IEEEcompsocitemizethanks{
\IEEEcompsocthanksitem Z. Kong and W. Luo are with the Shenzhen Campus of Sun Yat-sen University, Shenzhen, China, (e-mail: \{kongzhecn, whluo.china\}@gmail.com).
\IEEEcompsocthanksitem W. Zhang and Y. Li. are with Jarvis Research Center, Tencent YouTu Lab, (e-mail: zhangwentianml@gmail.com, leeyuexiang@163.com).
\IEEEcompsocthanksitem T. Wang is with the State Key Laboratory for Novel Software Technology, Nanjing University, Nanjing, China, (e-mail: taowangzj@gmail.com).
\IEEEcompsocthanksitem K. Zhang is with the Harbin Institute of Technology, Shenzhen, China, (e-mail: super.khzhang@gmail.com).
\IEEEcompsocthanksitem X. Tang is with the Department of Electronic and Electrical Engineering, Southern University of Science and Technology, Shenzhen, China, (e-mail: tangxy@sustech.edu.cn).
}
}


\maketitle

\begin{abstract}
Face recognition systems have raised concerns due to their vulnerability to different presentation attacks, and system security has become an increasingly critical concern.
Although many face anti-spoofing (FAS) methods perform well in intra-dataset scenarios, their generalization remains a challenge. To address this issue, some methods adopt domain adversarial training (DAT) to extract domain-invariant features. However, the competition between the encoder and the domain discriminator can cause the network to be difficult to train and converge. In this paper, we propose a domain adversarial attack (DAA) method to mitigate the training instability problem by adding perturbations to the input images, which makes them indistinguishable across domains and enables domain alignment.
Moreover, since models trained on limited data and types of attacks cannot generalize well to unknown attacks, we propose a dual perceptual and generative knowledge distillation framework for face anti-spoofing that utilizes pre-trained face-related models containing rich face priors.
Specifically, we adopt two different face-related models as teachers to transfer knowledge to the target student model. The pre-trained teacher models are not from the task of face anti-spoofing but from perceptual and generative tasks, respectively, which implicitly augment the data.
By combining both DAA and dual-teacher knowledge distillation, we develop a dual teacher knowledge distillation with domain alignment framework (DTDA) for face anti-spoofing.
The advantage of our proposed method has been verified through extensive ablation studies and comparison with state-of-the-art methods on public datasets across multiple protocols.
\end{abstract}


\section{Introduction}

Face recognition techniques \cite{deng2019arcface} have been widely used in authentication systems, such as mobile financial payment, access control, and other security scenarios. However, face recognition systems are vulnerable to different types of attacks, including 3D mask attacks, print attacks, and replay attacks. Thus the security of these systems has gradually become an increasingly critical concern. 

\begin{figure}[t]
  \centering
  \includegraphics[width=.47\textwidth]{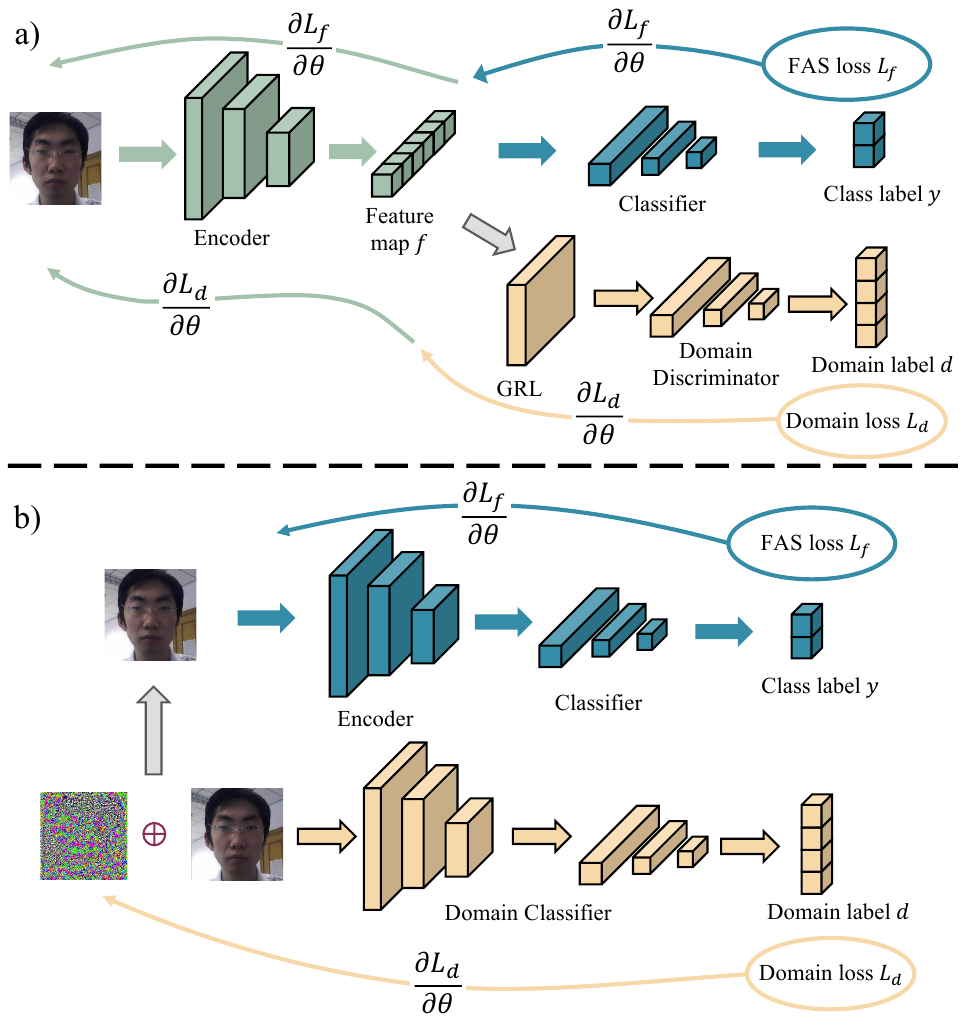}
  \caption{\textbf{Comparison of domain adversarial training (DAT) and the proposed domain adversarial attack (DAA) method.} a) is the DAT method. After the feature encoder, a gradient reversal layer (GRL) is inserted. A feature encoder is trained for producing the domain-invariant features, competing with domain discriminators simultaneously during the learning process, which gradually guides the learned features to be domain indistinguishable for face anti-spoofing. 
  b) is our proposed DAA method. A domain classifier is adopted to generate perturbations aiming at making the input image domains indistinguishable, which can encourage the model to learn domain-invariant features for face anti-spoofing and improve its generalization to the target domain.
  }
  \label{fig:motivation}
\end{figure}

To protect face recognition systems from attacks, there is a growing interest in investigating face anti-spoofing (FAS) methods to distinguish between live and spoof faces.
These methods are capable of extracting distinctive features, including handcrafted features \cite{de2013can,yang2013face,patel2016secure,boulkenafet2016face} and deep features \cite{wang2020cross,yu2020searching}, and are formulated as a binary classification problem.
Although previous detectors \cite{maatta2011face,wen2015face,Liu_2018_CVPR} perform well under intra-database testing scenarios, the distribution of testing samples in practice may differ from that of training samples, especially when the attack type is unseen. 
Discrepancies between training and testing data can lead to inferior detector performance, indicating a need to improve its generalization ability.
To address this problem, domain generalization (DG) \cite{maatta2011face,Wang_2022_CVPR,shao2019multi,qin2020learning,shao2020regularized,guo2022multi,sun2023rethinking,liu2023towards,yan2022domain} methods are proposed to minimize the distribution discrepancy between the source and target domains. 
Some face anti-spoofing methods \cite{Jia_2020_CVPR,Wang_2022_CVPR} leverage Domain Adversarial Training (DAT) \cite{ganin2015unsupervised} for domain alignment, as shown in Fig. \ref{fig:motivation} (a).
Specifically, a feature generator is trained to compete with a domain discriminator to make the features of different domains indistinguishable. Despite its power for extracting domain-invariant features, DAT is known to be challenging to train and converge. Since the detrimental and biased rivalry between the encoder and the domain discriminator, the training is unstable \cite{zhang2023free}.
It is crucial to develop an effective algorithm to address both domain alignment and unstable DAT training. Adversarial attack is a powerful technique that can influence the attention region in an image, making it possible to restrain domain-specific regions within the image.
Therefore, we propose to use adversarial attacks to achieve domain alignment as shown in Fig. \ref{fig:motivation} (b).

In addition, in machine learning, sufficient labeled data is crucial to improve the model generalization ability. 
However, for face anti-spoofing, it is expensive to collect large-scale labeled data. 
It is well known that knowledge distillation \cite{hinton2015distilling} is effective for transferring knowledge from teacher to student.
Therefore, to solve this problem, we attempt to mine additional data from other pre-trained models using knowledge distillation to achieve the goal of data augmenting in an implicit way.

To this end, we propose a dual teacher knowledge distillation with domain alignment framework, denoted as \textbf{DTDA} for face anti-spoofing, as depicted in Fig. \ref{fig:framework}. 
Our approach employs a domain classifier to generate perturbations for input images, making images from different domains indistinguishable and forming a domain adversarial attack method (DAA). This encourages the model to learn domain-invariant features for face anti-spoofing and improves its generalization to the target domain. 
To address the challenge of insufficient large-scale face data, we adopt knowledge distillation to acquire face priors. Specifically, we train our student network under the guidance of teacher models and propose dual-teacher models from perceptual and generative tasks, respectively.
The dual teacher models contain rich face priors implied in the employed a large amount of training face data.
Exploiting the strengths of various teacher models can bring significant advantages, with each teacher model being responsible for what they are proficient in and complimenting others.
Specifically, the dual teacher models adopted in our approach are tailored for face recognition and face attribute editing tasks, respectively.
Together, these face models, as dual teacher networks, provide rich, diverse, and useful face-prior knowledge to the student model via knowledge distillation, which achieves data augmentation in an implicit way.
By combining DAA and dual-teacher knowledge distillation, we further enhance the generalization potential of the student model.
During training, the inputs to both the student and the teacher models are adversarial samples.
Moreover, we train face recognition, face attribute editing, and face anti-spoofing tasks simultaneously in a multi-task learning manner.
As a result, the labeled data from all tasks can be aggregated to alleviate the problem of insufficient data. Our contributions are summarized as follows:

\begin{itemize}
\item
We propose a domain adversarial attack method to achieve domain alignment. Generating perturbations by leveraging the gradient from the domain classifier, we make the original images indistinguishable across domains. Utilizing domain-specific data, our model can extract domain-invariant features for FAS, which improves the model's generalization on the target domain.
\item 
We propose a dual-teacher knowledge distillation method using perceptual and generative models to achieve data augment in an implicit way. Without introducing additional network parameters, our model acquires diverse and rich facial priors from pre-trained models that are freely available. By leveraging these facial priors, we alleviate the problem of insufficient data, which boosts the generalization of the model.
\item 
Experimental results on public datasets under the cross-dataset testing settings verify the generalization of our proposed method. Specifically, when trained on OULU-NPU, MSU-MFSD, and Idiap Replay-Attack datasets and evaluated on CASIA-FASD, the proposed method achieves $6.67\%$ in terms of HTER, outperforming the state-of-the-art methods.
\end{itemize}

\section{Related Work}


\subsection{Face Anti-spoofing}

Existing face anti-spoofing methods can be broadly categorized as traditional and deep learning-based methods. 
Traditional methods typically rely on hand-craft features such as LBP \cite{de2013can}, HOG \cite{yang2013face}, SURF \cite{patel2016secure} and Surf \cite{boulkenafet2016face} for face anti-spoofing. However, these methods heavily rely on expert knowledge, and their performance may not be robust enough. 
Deep learning-based methods use neural networks to adaptively extract features \cite{liu2022fingerprint,Liu_2018_CVPR,Liu_2018_ECCV,liu20163d,yu2020face,yang2019face,zhang2022effective,wu2021dual,arashloo2020unseen}. 
However, these data-driven methods rely on data and thus exhibit the generalization issue. Specifically, while these models can achieve satisfactory results in the training distribution, their performance may degrade under shift distributions caused by different sensors, localization, and illumination conditions \cite{shao2019multi,wang2020cross}.  
To address this issue, specific network architectures \cite{yu2020searching,yu2020fas} are designed to extract robust features to strengthen the generalization. Some methods \cite{liu2020disentangling,zhang2020face,wang2020cross} disentangle face images into liveness features and content features, which can improve the generalization by focusing more attention on the liveness feature. Jia \etal \cite{Jia_2020_CVPR} design a single-side adversarial loss to align features of different domains and propose an asymmetric triplet loss to separate the spoof faces of different domains and aggregate the real ones. 
Wang \etal \cite{Wang_2022_CVPR} propose a Shuffled Style Assembly Network (SSAN) to extract and reassemble different content and style features for face images. Meta-learning-based methods \cite{chen2021generalizable,qin2021meta,shao2020regularized,qin2020learning,wang2021self} are also introduced for regular optimization. Self-supervised learning-based methods \cite{10051654,wang2022patchnet,wang2023consistency} extract face-related semantic features in pretext tasks, which are useful for face anti-spoofing.

In contrast to previous methods, our study employs knowledge distillation to enhance the generalization of face anti-spoofing with face recognition and editing networks. Our model is trained under the guidance of soft labels generated by teacher models. Compared with hard labels, soft labels can mine more intra-class variance and inter-class distance information, providing better supervision.

\subsection{Adversarial Attack}
Szegedy \etal \cite{szegedy2013intriguing} find that adding a quasi-imperceptible, carefully crafted perturbation to an image can mislead the deep neural network to make a wrong decision. Numerous effective adversarial attack methods, such as Projected Gradient Descent Attack (PGD) \cite{madry2017towards}, Gradient Sign Method (FGSM) \cite{goodfellow2014explaining}, Carilini and Wagner Attack (CW) \cite{carlini2017towards}, and MI-FGSM \cite{dong2018boosting} have been proposed. The category for adversarial attacks can be roughly divided into white-box attacks and black-box attacks. White-box attacks are allowed to access all the parameters and concrete structure information of the attacked model when generating adversarial examples, and black-box attacks are to know only part of the attacked model’s output when generating adversarial examples. 

In the proposed method, we use adversarial attacks not to fool the network into distinguishing live and spoof images, but to suppress the domain-specific regions in an image.
Our method aims to force the network to extract domain-invariant features that generalize well to the target domain.
We use the PGD method for white-box adversarial example generation due to its flexibility and effectiveness in conducting extensive experiments.

\subsection{Knowledge Distillation}
Knowledge distillation is introduced in \cite{hinton2015distilling}, which can transfer the knowledge of teacher models to the student model. As the teacher model's feature has a stronger representation power, its key strategy is to make students imitate the teacher's output as close as possible. The student is supervised by the hard ground-truth labels and soft labels from the teacher's last linear layer.
Due to its strong performance, knowledge distillation has been widely used in practical deployment scenarios combined with network pruning and model quantization. Recently, numerous methods study \cite{zagoruyko2016paying,park2019relational,tung2019similarity,romero2014fitnets,tian2019contrastive,zhao2023learn,zhang2023unsupervised} distill the knowledge in the intermediate features. It is proposed in \cite{zagoruyko2016paying} to mimic the attention maps in convolutional layers of the teacher model. In \cite{heo2019comprehensive}, a teacher transforms, and a new distance function is proposed to measure the distance for distillation. 
Some studies have explored face anti-spoofing through knowledge distillation.
Existing methods \cite{qin2021meta,li2020face,li2022one,liu2022cross} attempt to improve the generalization for face anti-spoofing by training a strong teacher model.
Qin \etal \cite{qin2021meta} train a meta-teacher in a bi-level optimization manner learning to supervise the PA detectors to learn rich spoofing cues, and the meta-teacher is explicitly trained to better teach the detector.
\cite{li2022one} is a one-class method that utilizes a few genuine face samples of the target domain to help the teacher model extract more discriminative features and then distill the student model in the source domain. 
In \cite{li2020face}, data from a richer and related domain are leveraged to train the teacher model. 
In order to achieve higher performance, \cite{liu2022cross} directly adopts a Transformer network \cite{dosovitskiy2020image} as the teacher model. 
However, the performance of the student model is limited if the teacher model cannot perform well. 
Previous knowledge distillation methods for face anti-spoofing train a single teacher model using only the face anti-spoofing task. In contrast, our proposed method differs significantly from existing techniques in two ways.
Firstly, we propose a domain adversarial attack method to achieve domain alignment.
By using perturbations generated by the domain classifier, we make the input image domains indistinguishable, forcing the model to learn domain-invariant features.
Secondly, we propose a dual-teacher knowledge distillation method for face anti-spoofing. 
Specifically,  the teacher models in our proposed method are trained not from the task of face anti-spoofing, but from perceptual and generative tasks, providing the student model with abundant face priors. Additionally, by training multiple face tasks simultaneously, our method alleviates the data insufficiency problem and improves the generalization of the student model.

\begin{figure*}[ht]
  \centering
  \includegraphics[width=.98\textwidth]{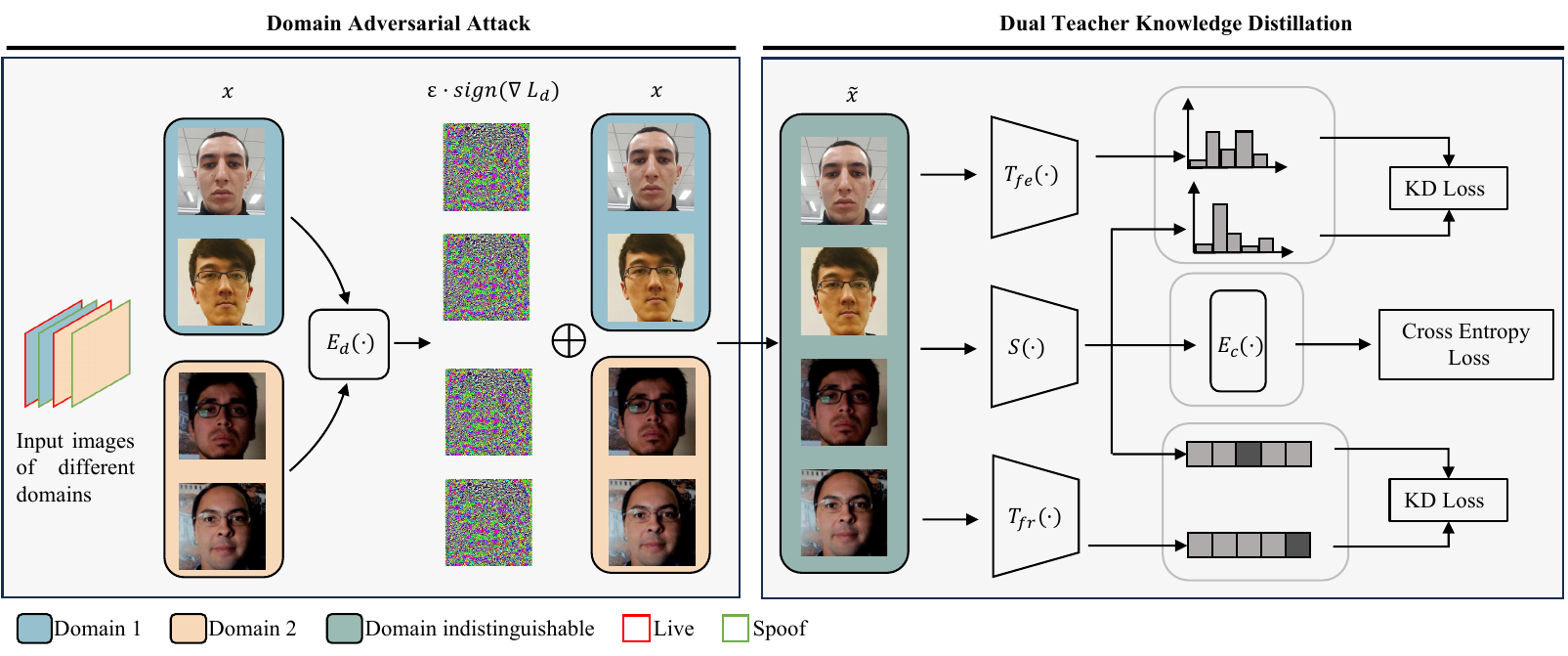}
  \caption{\textbf{The overall architecture of our face anti-spoofing via dual teachers with domain alignment (DTDA) method.} The input faces with different color backgrounds represent examples of different domains. During the adversarial attack stage, a domain classifier is utilized to generate perturbations that make the input image domains indistinguishable, thereby forcing the model to learn domain-invariant features. In the dual teacher knowledge distillation stage, two teacher models transfer face priors to the student model. The dual teacher models are sourced from perceptual and generative tasks trained on millions of face data, further improving the generalization of the model.
  }
  \label{fig:framework}
\end{figure*}
\section{Proposed Method}
As shown in Fig. \ref{fig:framework}, the proposed DTDA method contains two stages during training. 
Firstly, the domain adversarial attack method is leveraged to generate domain indistinguishable images.
Given a batch of training data containing live and spoof images from different domains, we derive the corresponding perturbations using a pre-trained domain classifier. 
By adding the perturbations to the input images, we can make the adversarial image domains indistinguishable, allowing the model to learn domain-invariant features that improve the model's generalization ability.
Different facial tasks share the face priors and they may benefit from each other.
The face recognition model and face attribute editing model are trained on large-scale face datasets, which capture rich texture and shape priors of the face. 
And then, we leverage knowledge distillation to exploit common priors and enhance the generalization capability of our model. Adversarial images are used to train both the student and teacher models during the knowledge distillation stage.
Our proposed method incorporates two teacher models: a face recognition teacher $T_{fr}$ and a face attribute editing teacher $T_{fa}$. 
Both the student and teacher models use adversarial samples generated by our domain adversarial attack method as input. 
The teacher models produce soft labels for face recognition and face attribute editing, respectively, while hard labels for face anti-spoofing are derived from annotations. 
Hence, under the guidance of both soft and hard labels, the student model benefits from richer supervision. 
We train face recognition, face attribute editing, and face anti-spoofing tasks in a multi-task learning manner.
This enables the knowledge contained in one task to be leveraged by others, resulting in improved generalization performance for all tasks.

\subsection{Domain Adversarial Attack}
In real-world scenarios, face images are often captured by different cameras under varying illumination, resolution, and backgrounds.
This leads to a domain shift between the source and target domains, making domain alignment essential.
Domain adversarial training is a popular approach for mitigating the distribution discrepancies between source and target domains.
However, the competition between the encoder and the domain discriminator in domain adversarial learning can be unstable, making the network difficult to train and converge.
To address this issue, we propose a domain adversarial attack method that adds perturbations to the input images.
This makes the input image domains indistinguishable, achieving domain alignment and mitigating the instability problem commonly encountered in domain adversarial training.

Given live and spoof input images from different domains, we first train a domain classifier $E_d(\cdot)$, which is trained with a cross-entropy loss defined as follows
\begin{equation}
    \label{eq:l-domain}
    \begin{aligned}
        \mathcal{L}_{d} = -\frac{1}{N} \sum_{}^{N}[dlog(\hat{p}_{d})+(1-d)log(1-\hat{p}_{d})],
    \end{aligned}
\end{equation}
where $d$ is the domain label and $\hat{p}_{d}$ is the predicted result of input images $x$ .

To make the input image domain indistinguishable, we use an adversarial attack to force the network to produce incorrect predictions, which is formulated as
\begin{equation}
    \label{eq:daa}
    \begin{aligned}
        \widetilde{x} = x + \varepsilon sign(\nabla_{x}\mathcal{L}_{d}).
    \end{aligned}
\end{equation}

To ensure the effectiveness of the domain attack during distillation, the network architectures of the domain classifier and student are similar.
By using the domain adversarial attack, the domains of input images are indistinguishable, allowing the network to exploit more common discriminative cues for face anti-spoofing.

\subsection{Knowledge Distillation for Face Anti-spoofing}
Face anti-spoofing models are susceptible to illumination variations, and training them with sufficient data and annotation is challenging.
Collecting large-scale data and annotations is expensive and time-consuming, and privacy and fairness issues make it difficult or even impossible to collect data under all conditions.
To alleviate the shortage of data volume, we aim to leverage face priors from other pre-trained models through knowledge distillation.
The knowledge distillation framework is depicted in the right part of Fig. \ref{fig:framework}, which consists of a student network and two teacher networks. One teacher network, denoted as $T_{fr}$ is designed for face recognition, while the other, denoted as $T_{fa}$, is specific for face attribute editing. 
Both teacher networks are pre-trained on large-scale face datasets. 
With the guidance of $T_{fa}$ and $T_{fr}$ in knowledge distillation, multiple tasks are trained simultaneously. By sharing face priors, the model can improve generalization performance.

We use $x$ and $y$ to represent the input face image and its corresponding label. $\widetilde{x}$ is the adversarial sample generated from $x$.
We input $\widetilde{x}$ into both the teacher and student models to extract domain-invariant features. Therefore, the proposed method forces the model to learn a generalized feature space for images from different domains.
The outputs produced by the dual teacher models are represented as:
\begin{equation}
    \label{eq:fa-out}
    \begin{aligned}
        p_{edi} = T_{fa}(\widetilde{x}),
    \end{aligned}
\end{equation}
\begin{equation}
    \label{eq:fr-out}
    \begin{aligned}
        p_{rec} = T_{fr}(\widetilde{x}).
    \end{aligned}
\end{equation}
Here, $p_{edi}$ is the output of the face attribute editing teacher, and $p_{rec}$ is the output of face recognition teacher $T_{fr}$.
We adopt knowledge distillation in the student model, which shares the same network backbone for each face task but with different embedding layers to match the corresponding output of the teacher model.
The student model comprises a feature encoder $S$ and three parallel embedding layers, editing embedding $E_{e}$, recognition embedding $E_{r}$, and classification embedding $E_{c}$. 
Specifically, given an input image $x$, a shared feature map can be encoded by
\begin{equation}
    \label{eq:s-out}
    \begin{aligned}
        f = S(x).
    \end{aligned}
\end{equation}
The corresponding outputs can be calculated from $f$. For instance, $E_{e}(f)$ generates the output $\hat{p}_{edi}$, while $E_{r}$ and $E_{c}$ generate $\hat{p}_{rec}$ and $\hat{p}_{fas}$, respectively. 
For the face attribute editing task, the teacher model generates the ground truth labels. We aim to maximize the consistency of logit-based class probability distributions between the output of student and face attribute editing teacher via knowledge distillation (KD) loss,
\begin{equation}
    \label{eq:l-fa}
    \begin{aligned}
        \mathcal{L}_{fa} = 
        \mathcal{KL}(softmax(\frac{p_{edi}}{\tau}), softmax(softmax(\frac{\hat{p}_{edi}}{\tau})), \\
    \end{aligned}
\end{equation} 
where $\mathcal{KL}$ indicates the Kullback-Leibler divergence, which can measure the distance between the categorical probability distribution of the student and teacher models. The temperature hyper-parameter $\tau$ is used in the softmax function.
In addition to the KD loss used for face attribute editing, we also use it to optimize $S$ and $E_{r}$ for face recognition:
\begin{equation}
    \label{eq:l-fr}
    \begin{aligned}
        \mathcal{L}_{fr} = \mathcal{KL}(softmax(\frac{p_{rec}}{\tau}), softmax(softmax(\frac{\hat{p}_{rec}}{\tau})).
    \end{aligned}
\end{equation}
Face recognition and face attribute editing are auxiliary tasks and can extract additional face priors for face anti-spoofing. To optimize our model for face anti-spoofing, we adopt the cross-entropy loss as:
\begin{equation}
    \label{eq:l-fas}
    \begin{aligned}
        \mathcal{L}_{f} = -\frac{1}{N} \sum_{}^{N}[p_{fas}log(\hat{p}_{fas})+(1-p_{fas})log(1-\hat{p}_{fas})].
    \end{aligned}
\end{equation}
To provide a clear view of the proposed method, we illustrate our algorithm in Alg. \ref{alg:dpgkd}.

\begin{algorithm}[tbp]
    \caption{Face Anti-Spoofing using DTDA}
    \label{alg:dpgkd}
    \textbf{Date}: Training data $\mathcal{X}_t$\\
    \textbf{Init}: Domain Classifier $E_{d}(\cdot)$; \\
        \hspace*{0.26in} Pre-trained Face Recognition Teacher $T_{fr}(\cdot)$; \\
        \hspace*{0.26in} Pre-trained Face Attribute Editing Teacher $T_{fe}(\cdot)$; \\
        \hspace*{0.26in} Student $S(\cdot)$; \\
        \hspace*{0.26in} Editing Embedding $E_{e}(\cdot)$; \\
        \hspace*{0.26in} Recognition Embedding $E_{r}(\cdot)$; \\
        \hspace*{0.26in} Classification Embedding $E_{c}(\cdot)$; \\
    \textbf{Output}: Model parameters $S(\cdot)$, $E_{e}(\cdot)$, $E_{r}(\cdot)$ and $E_{c}(\cdot)$;
    \begin{algorithmic}[1]
        \WHILE{Model parameters have not converged}
        \FOR{$x$ in $\mathcal{X}_t$}
            \STATE Generate domain indistinguishable sample $\widetilde{x}$ through $x + \varepsilon sign(\nabla_{x}\mathcal{L}_{domain})$;\\
            \STATE Extract feature $f$ through $S(\widetilde{x})$;\\
            \STATE Derive embedding $\hat{p}_{edi}$ through $E_{e}(f)$;\\
            \STATE Obtain embeddings $\hat{p}_{rec}$, $\hat{p}_{fas}$ through $E_{r}(f)$ and $E_{c}(f)$;\\
            \STATE Calculate $\mathcal{L}_{fas}$ through Eq. \eqref{eq:l-fas}; \\
            \STATE Derive class probability $p_{edi}$ through $T_{fa}(\widetilde{x})$;\\
            \STATE Obtain face recognition features $p_{rec}$ from $T_{fr}(\widetilde{x})$;\\   
            \STATE Calculate $\mathcal{L}_{fa}$ and $\mathcal{L}_{fr}$ through Eq. \eqref{eq:l-fa}, \eqref{eq:l-fr};\\
            \STATE Update $S(\cdot)$, $E_{e}(\cdot)$, $E_{c}(\cdot)$ and $E_{c}(\cdot)$ by minimizing Eq.~\eqref{eq:l-sum};\\
        \ENDFOR
        \ENDWHILE
    \end{algorithmic}
\end{algorithm}

\subsection{Loss Function}

Multi-task learning has been shown to improve generalization \cite{zhang2021survey}, so we train the face recognition, face attribute editing, and face anti-spoofing tasks simultaneously to alleviate overfitting.
During training, we freeze the network parameters of the teacher models and optimize only the student network and its corresponding embedding layers.

To summarize, the total loss function $\mathcal{L}_{sum}$ for face anti-spoofing is defined as:
\begin{equation}
    \label{eq:l-sum}
    \begin{aligned}
        \mathcal{L}_{sum} = \mathcal{L}_{f} + 
        \lambda_{1}\mathcal{L}_{fr} + 
        \lambda_{2}\mathcal{L}_{fa} ,
    \end{aligned}
\end{equation}
where $\lambda_{1}$ and $\lambda_{2}$ are hyper-parameters to balance the proportion of different loss terms.

\section{Experiment}

To explore the effectiveness of our proposed method, we evaluate DTDA on numerous publicly available datasets. In the following, we first introduce the dataset and experimental protocol. Then, we validate the effectiveness of the proposed method by conducting ablation studies. Finally, we compare DTDA with state-of-the-art methods to verify its superiority.

\subsection{Datasets and Protocols}

\textbf{Datasets}. We use four publicly available datasets to evaluate the performance of the proposed method, including OULU-NPU \cite{boulkenafet2017oulu} (denoted as O), CASIA-FASD \cite{zhang2012face} (denoted as C), Idiap Replay-Attack \cite{chingovska2012effectiveness} (denoted as I), MSU-MFSD \cite{wen2015face} (denoted as M) and CelebA-Spoof \cite{zhang2020celeba} (denoted as S).  

\begin{table}[tbp]
\centering
\caption{Summary of the face anti-spoofing datasets.}
\label{datasets}
\setlength\tabcolsep{4pt}
\resizebox{.98\columnwidth}{!}{%
\begin{tabular}{c|c|cc|cc}
\hline
\multirow{2}{*}{Dataset} & \multirow{2}{*}{Attack type}                                                           & \multicolumn{2}{c|}{Subject}      & \multicolumn{2}{c}{Video}          \\ \cline{3-6} 
                         &                                                                                        & \multicolumn{1}{c|}{train} & test & \multicolumn{1}{c|}{real} & attack \\ \hline
Oulu-NPU (O)             & \begin{tabular}[c]{@{}c@{}}Printed phone\\ Display phone\\ Replayed video\end{tabular} & \multicolumn{1}{c|}{35}    & 20   & \multicolumn{1}{c|}{1980} & 3960   \\ \hline
CASIA-FASD (C)           & \begin{tabular}[c]{@{}c@{}}Printed phone\\ Cut phone\\ Replayed video\end{tabular}     & \multicolumn{1}{c|}{20}    & 30   & \multicolumn{1}{c|}{150}  & 450    \\ \hline
Idiap Replay-Attack (I)  & \begin{tabular}[c]{@{}c@{}}Printed photo\\ Replayed video\end{tabular}                 & \multicolumn{1}{c|}{30}    & 20   & \multicolumn{1}{c|}{390}  & 640    \\ \hline
MSU-MFSD (M)             & \begin{tabular}[c]{@{}c@{}}Printed photo\\ Cut phone\\ Replayed video\end{tabular}     & \multicolumn{1}{c|}{18}    & 17   & \multicolumn{1}{c|}{110}  & 330    \\ \hline
\end{tabular}}
\end{table}

To evaluate the effectiveness of our method, we follow \cite{Jia_2020_CVPR} and use two protocols for cross-dataset testing. 
In cross-dataset testing, we conduct experiments on four datasets: OULU-NPU \cite{boulkenafet2017oulu}, CASIA-FASD \cite{zhang2012face}, Idiap Replay-Attack \cite{chingovska2012effectiveness}, MSU-MFSD \cite{wen2015face}.
The fine-grained subject number, video number, and attack types for cross-dataset testing are summarized in Table \ref{datasets}.
In Protocol-\uppercase\expandafter{\romannumeral1}, three datasets are used as training sets, and the remaining one is used for testing. Thus, the experiment consists of four groups in total: O\&C\&I to M, O\&M\&I to C, O\&C\&M to I, and I\&C\&M to O. In Protocol-\uppercase\expandafter{\romannumeral2}, we use two datasets for training and one for testing. Specifically, we have two testing tasks in total: M\&I to C and M\&I to O.

For intra-dataset testing, we performed experiments on CelebA-Spoof \cite{zhang2020celeba} dataset. This dataset is a comprehensive face anti-spoofing collection, encompassing 625,537 images of 10,177 subjects. The dataset incorporates 43 detailed attributes covering aspects such as facial features, illumination, environment, and various spoof types.
Furthermore, each image within the CelebA-Spoof dataset is meticulously annotated with the 43 attributes. This abundance of annotation significantly enhances the dataset's diversity, providing a more illustrative and comprehensive foundation for face anti-spoofing research.

\begin{table*}[ht]
\small
\centering
\caption{Results of ablation studies in terms of HTER (\%) and AUC (\%) under the cross-dataset setting on OULU-NPU (O), CASIA-FASD (C), Idiap Replay-Attack (I) and MSU-MFSD (M). Among these models, the perceptual model is derived from the face recognition task, while the generative model corresponds to the face attribute editing task.}
\label{tab:ablation_study}
\setlength\tabcolsep{4pt}
\resizebox{2\columnwidth}{!}{%
\begin{tabular}{cccc|cc|cc|cc|cc}
\hline
\multirow{2}{*}{Baseline} & \multirow{2}{*}{\begin{tabular}[c]{@{}c@{}} Perceptual (Face\\ Recognition)\end{tabular}} & \multirow{2}{*}{\begin{tabular}[c]{@{}c@{}}Generative (Face \\Attribute Editing)\end{tabular}} & \multirow{2}{*}{\begin{tabular}[c]{@{}c@{}}Domain \\ Adversarial Attack\end{tabular}} & \multicolumn{2}{c|}{O\&C\&I to M} & \multicolumn{2}{c|}{O\&M\&I to C} & \multicolumn{2}{c|}{O\&C\&M to I} & \multicolumn{2}{c}{I\&C\&M to O} \\ \cline{5-12} 
                          &                                                                             &                                                                                   &                                                                                       & HTER            & AUC             & HTER            & AUC             & HTER            & AUC             & HTER            & AUC            \\ \hline
$\surd$                   & $\times$                                                                    & $\times$                                                                          & $\times$                                                                              & 11.43           & 95.07           & 16.00           & 91.45           & 24.74           & 77.25           & 17.08           & 90.94          \\ \hline
$\surd$                   & $\surd$                                                                     & $\times$                                                                          & $\times$                                                                              & 8.57            & 96.58           & 13.33           & 93.16           & 19.90           & 75.22           & 14.45           & 91.84          \\ \hline
$\surd$                   & $\times$                                                                    & $\surd$                                                                           & $\times$                                                                              & 8.57            & 95.14           & 11.22           & 94.78           & 20.81           & 75.11           & 16.39           & 90.82          \\ \hline
$\surd$                   & $\surd$                                                                     & $\surd$                                                                           & $\times$                                                                              & 7.14            & 97.20           & 9.33            & 95.62           & 13.76           & \textbf{92.55}           & 13.83           & 92.69          \\ \hline
\rowcolor[HTML]{EFEFEF}
$\surd$                   & $\surd$                                                                     & $\surd$                                                                           & $\surd$                                                                               & \textbf{5.71}   & \textbf{98.03}  & \textbf{6.67}   & \textbf{97.27}  & \textbf{13.12}  & {92.24}  & \textbf{13.13}  & \textbf{94.24} \\ \hline
\end{tabular}%
}
\end{table*}

\begin{figure*}[ht]
  \centering
  \includegraphics[width=0.75\textwidth]{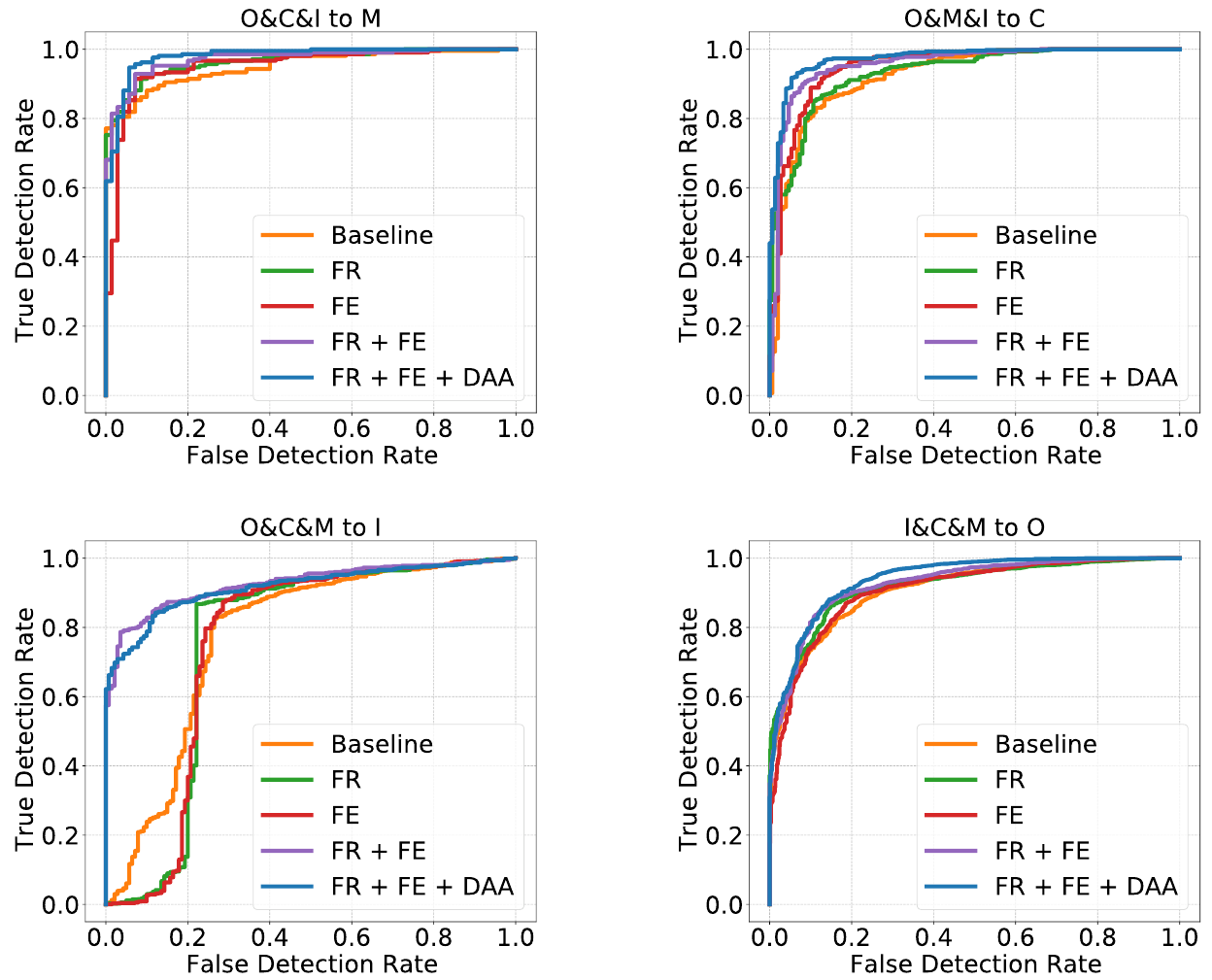}
  \caption{\textbf{ROC curves for the ablation study under the cross-dataset testing.} FR denotes the face recognition network, FE denotes the face attribute editing network and DAA denotes the domain adversarial attack method.
  }
  \label{fig:roc}
\end{figure*}

\textbf{Implementation Details}. In our experiment, the MTCNN algorithm \cite{zhang2016joint} is adopted for face detection and alignment. All the detected faces are resized to $256\times256$. During training, we randomly select one frame from each video. ResNet18 \cite{he2016deep} is set as the backbone for the student model. For face recognition, we use the pre-trained model from ArcFace \cite{deng2019arcface}. And we only use the pre-trained discriminator of StarGAN \cite{choi2018stargan} for face attribute editing. 
The SGD optimizer with a momentum of $0.9$ and weight decay of 5e-4 is used for the optimization. The learning rate is $0.1$, and the batch size for training is $96$.
Our framework is implemented using PyTorch, and all the experiments are carried out using NVIDIA Tesla V100 and RTX 3090 GPUs.

\textbf{Performance Metrics}. Following the work of \cite{Jia_2020_CVPR}, we use the Half Total Error Rate (HTER) and the Area Under Curve (AUC) as the evaluation metrics for cross-dataset testing for a fair comparison.
In intra-dataset testing, we follow the original protocols and use FPR@Recall, AUC, Attack Presentation Classification Error Rate (APCER), Bona Fide Presentation Classification Error Rate (BPCER), and ACER metrics for a fair comparison. Notably, ACER is defined as follows:

\begin{equation}
    \label{eq:acer}
    \begin{aligned}
        ACER = \frac{APCER + BPCER}{2}
    \end{aligned}
\end{equation}

\subsection{Ablation Study}

\begin{table*}[tbp]
\caption{Performance comparison between the proposed method and the state-of-the-art methods in terms of HTER(\%) and AUC (\%) under the cross-dataset setting (Protocol-\uppercase\expandafter{\romannumeral1}).}
\centering
\label{tab:compare3to1}
\setlength\tabcolsep{8pt}
\resizebox{.98\textwidth}{!}{%
\begin{tabular}{c|cc|cc|cc|cc|cc}
\hline
\multirow{2}{*}{Method} & \multicolumn{2}{c|}{O\&C\&I to M} & \multicolumn{2}{c|}{O\&M\&I to C} & \multicolumn{2}{c|}{O\&C\&M to I} & \multicolumn{2}{c|}{I\&C\&M to O} & \multicolumn{2}{c}{Mean}       \\ \cline{2-11} 
                        & HTER            & AUC             & HTER            & AUC             & HTER            & AUC             & HTER            & AUC             & HTER          & AUC            \\ \hline
Auxiliary \cite{Liu_2018_CVPR}              & 22.72           & 85.88           & 33.52           & 73.15           & 29.14           & 71.69           & 30.17           & 77.61           & 28.89         & 77.08          \\
PAD-GAN \cite{wang2020cross}                & 17.02           & 90.10           & 19.68           & 87.43           & 20.87           & 86.72           & 25.02           & 81.47           & 20.65         & 86.43          \\
RFM \cite{shao2020regularized}                    & 13.89           & 93.98           & 20.27           & 88.16           & 17.30           & 90.48           & 16.45           & 91.16           & 16.98         & 90.95          \\
NAS-FAS \cite{yu2020fas}                & 16.85           & 90.42           & 15.21           & 92.64           & 11.63           & \textbf{96.98}  & 13.16           & 94.18           & 14.21         & 93.56          \\
EPCR \cite{wang2023consistency}                   & 12.50           & 95.30           & 18.90           & 89.70           & 14.00           & 92.40           & 17.90           & 90.90           & 15.83         & 92.08          \\
DRDG \cite{liu2021dual}                   & 12.43           & 95.81           & 19.05           & 88.79           & 15.56           & 91.79           & 15.63           & 91.75           & 15.67         & 92.04          \\
VLAD-VSA-M \cite{wang2021vlad}             & 11.43           & 96.44           & 20.79           & 86.32           & 12.29           & 92.95           & 21.20           & 86.93           & 16.43         & 90.66          \\
ANRL \cite{liu2021adaptive}                   & 10.83           & 96.75           & 17.83           & 89.26           & 16.03           & 91.04           & 15.67           & 91.90           & 15.09         & 92.24          \\
SSDG \cite{Jia_2020_CVPR}                 & 7.38            & 97.17           & 10.44           & 95.94           & 11.71           & 96.59           & 15.61           & 91.54           & 11.29         & 95.31          \\
SSAN-M \cite{Wang_2022_CVPR}                 & 10.42           & 94.76           & 16.47           & 90.81           & 14.00           & 94.58           & 19.51           & 88.17           & 15.10         & 92.08          \\
DF-DM \cite{10051654}                  & 7.14            & 97.09           & 15.33           & 91.41           & 14.03           & 94.30           & 16.68           & 91.85           & 13.30         & 93.66          \\
GDA \cite{zhou2022generative}                    & 9.20            & 98.00           & 12.20           & 93.00           & \textbf{10.00}  & 96.00           & 14.40           & 92.60           & 11.45         & 94.90          \\ \hline
\rowcolor[HTML]{EFEFEF}
Ours                    & \textbf{5.71}   & \textbf{98.03}  & \textbf{6.67}   & \textbf{97.27}  & 13.12           & 92.24           & \textbf{13.13}  & \textbf{94.24}  & \textbf{9.66} & \textbf{95.45} \\ \hline
\end{tabular}%
}
\end{table*}

To investigate the effectiveness of different components in our model, we conduct an ablation study using cross-dataset testing. We have five variants as the following.

\textbf{Variant1}: The first variant is denoted as \texttt{Baseline}, using a ResNet-18 pre-trained on ImageNet as the baseline method. After the features pass through the classifier, we only calculate the loss $\mathcal{L}_{fas}$ using the cross-entropy function, as shown in Eq. \eqref{eq:l-fas}.

\textbf{Variant2}: The second variant is \texttt{Baseline + Face Recognition}. In this experiment, face recognition serves as a teacher model for learning perceptual priors. 
To measure the output distance between the student and teacher models, we use the $\mathbf{KL}$ loss, similar to previous knowledge distillation methods \cite{hinton2015distilling}. 
The loss function for this variant, $\mathcal{L}_{fr}$, is shown in Eq. \eqref{eq:l-fr}. 
Therefore, the sum loss for this variant is $\mathcal{L}_{fas} + \lambda_1 \mathcal{L}_{fr}$, where $\lambda_1$ is a hyperparameter that balances the two loss terms.

\textbf{Variant3}: The third variant, \texttt{Baseline + Face Attribute Editing}, employs face attribute editing as another teacher model for learning generative priors. Similar to the face recognition teacher model mentioned above, we calculate the loss between the face editing teacher model and the student model using Eq. \eqref{eq:l-fa}, denoted as $\mathcal{L}_{fa}$. Variant3 is trained using $\mathcal{L}_{fas} + \lambda_2 \mathcal{L}_{fa}$, where $\lambda_2$ balances the two loss terms.

\textbf{Variant4}: In the fourth variant, \texttt{Baseline + Face Recognition + Face Attribute Editing}, we combine the face recognition and face attribute editing teacher models. The method is trained using Eq. \eqref{eq:l-sum}, with clean images used as input for both the student and teachers.

\textbf{Full method}: The fifth and final variant, \texttt{Baseline + Face Recognition + Face Attribute Editing + Domain Adversarial Attack}, is our full method. Similar to Variant 4, it uses Eq. \eqref{eq:l-sum} for network optimization. However, adversarial images are used as input for both the student and teacher models.

The qualitative comparison results are shown in Table \ref{tab:ablation_study} and the corresponding ROC curves are shown in Fig. \ref{fig:roc}. The training set and testing set are from different domains, which can show the generalization ability and effectiveness of our method. Compared to the baseline, both the face recognition and attribute editing networks improve the generalization of face anti-spoofing. Specifically, adopting face recognition yields a $3.25\%$ increase in mean HTER, while face attribute editing improves the mean AUC from $88.68\%$ to $88.96\%$.
Using both face-related teachers simultaneously yields better results compared to using a single-teacher model. 
Finally, using both the dual-teacher model and domain adversarial attack simultaneously achieves the best performance in terms of HTER and AUC, indicating the benefits of both face priors and DAA in face anti-spoofing.

\begin{table}[t]
\caption{Performance comparison between the proposed method and the state-of-the-art methods in terms of HTER(\%) and AUC (\%) under the cross-dataset setting (Protocol-\uppercase\expandafter{\romannumeral2}).}
\label{tab:compare2to1}
\centering
\setlength\tabcolsep{8pt}
\resizebox{.98\columnwidth}{!}{%
\begin{tabular}{c|cc|cc}
\hline
\multirow{2}{*}{Method} & \multicolumn{2}{c|}{O\&I to C}  & \multicolumn{2}{c}{M\&I to O}   \\ \cline{2-5} 
                        & HTER           & AUC            & HTER           & AUC            \\ \hline
MS-LBP \cite{maatta2011face}                 & 51.16          & 52.09          & 43.63          & 58.07          \\
IDA \cite{wen2015face}                    & 45.16          & 58.80          & 54.52          & 42.17          \\
CT \cite{boulkenafet2016face}                     & 55.17          & 46.89          & 53.31          & 45.16          \\
LBP-TOP \cite{maatta2011face}                & 45.27          & 54.88          & 47.26          & 50.21          \\
MADDG \cite{shao2019multi}                  & 41.02          & 64.33          & 39.35          & 65.10          \\
DR-MD-Net \cite{wang2020cross}                 & 31.67          & 75.23          & 34.02          & 72.65          \\ 
SSDG-M \cite{Jia_2020_CVPR}                 & 31.89          & 71.29          & 36.01          & 66.88          \\ 
ANRL \cite{liu2021adaptive}                 & 31.06          & 71.12          & 30.73          & 74.10          \\ 
SSAN-M \cite{Wang_2022_CVPR}                 & 30.00          & 76.20          & 29.44          & 76.62          \\ \hline
\rowcolor[HTML]{EFEFEF}
\textbf{Ours}           & \textbf{22.67} & \textbf{82.43} & \textbf{28.22} & \textbf{79.44} \\ \hline
\end{tabular}%
}
\end{table}

\subsection{Cross-Dataset Testing}

\textbf{Experiment in Leave-One-Out Setting.} To further verify the generalization of the proposed method, we compare our method with the state-of-the-art methods in two protocols. Table \ref{tab:compare3to1} shows the comparison results on Protocol-\uppercase\expandafter{\romannumeral1}. In this protocol, we conduct cross-dataset testing using a leave-one-out strategy: three datasets are selected for training and the rest for testing. We compare our methods with the state-of-the-art methods, including Auxiliary \cite{Liu_2018_CVPR}, PAD-GAN \cite{wang2020cross}, RFM \cite{shao2020regularized}, SDA-FAS \cite{wang2021self}, NAS-FAS \cite{yu2020fas}, EPCR \cite{wang2023consistency}, DRDG \cite{liu2021dual}, VLAD-VSA-M \cite{wang2020cross}, ANRL \cite{liu2021adaptive}, SSDG \cite{Jia_2020_CVPR}, SSAN-M \cite{Wang_2022_CVPR}, DF-DM \cite{10051654} and GDA \cite{zhou2022generative}. The proposed method achieves the best performance in all settings except for the case of O\&C\&M to I, where the proposed method still obtains results on par with the SOTA methods.
Among all the cases, the most significant improvement is obtained on the setting O\&M\&I to C, which outperforms the state-of-the-art method by more than $3\%$ in terms of HTER. It further verifies the effectiveness and generalization capability of our proposed method.

\textbf{Experiment on Limited Training Data}. In Protocol-\uppercase\expandafter{\romannumeral2}, limited source data can be leveraged for training, which can further measure the generalization of the method. In this protocol, two datasets are selected for training, and we use another two datasets for testing respectively. Experiment results are shown in Table \ref{tab:compare2to1}. Our method achieves the best HTER and AUC under the limited training data, with a significant improvement over other methods. Although only two source domains are available, the proposed method can still push the distribution of live and spoof faces away, demonstrating the superiority of our method.

\begin{table*}[]
\centering
\caption{The obtained results are reported in terms of FPR@Recall (\%), AUC (\%), APCER (\%), BPCER (\%) and ACER (\%) under the intra-dataset setting on CelebA-Spoof dataset.}
\setlength\tabcolsep{8pt}
\label{tab:inta}
\resizebox{2\columnwidth}{!}{%
\begin{tabular}{c|c|ccccccc}
\hline
Protocol           & Sensors types & FPR=1\% & FPR=0.5\% & FPR=0.1\% & AUC        & APCER     & BPCER     & ACER      \\ \hline
1                  &               & 95.72   & 94.01     & 90.39     & 99.75      & 2.43      & 2.40      & 2.42      \\ \hline
\multirow{4}{*}{2} & Low           & -       & -         & -         & 100.00     & 0.09      & 0.10      & 0.10      \\
                   & Middle        & -       & -         & -         & 99.71      & 2.09      & 2.15      & 2.12      \\
                   & High          & -       & -         & -         & 99.97      & 0.50      & 0.49      & 0.50      \\ \cline{2-9} 
                   & Average       & -       & -         & -         & 99.89±0.16 & 0.89±1.06 & 0.91±1.09 & 0.90±1.07 \\ \hline
\end{tabular}%
}
\end{table*}

\begin{table*}[]
\centering
\caption{Performance comparison between the proposed method and the state-of-the-art methods in terms of FPR@Recall (\%), AUC (\%), APCER (\%), BPCER (\%) and ACER (\%) under the intra-dataset setting on CelebA-Spoof dataset.}
\setlength\tabcolsep{8pt}
\label{tab:intra-comp}
\resizebox{2\columnwidth}{!}{%
\begin{tabular}{c|c|ccccccc}
\hline
Protocol           & Methods          & FPR=1\%        & FPR=0.5\%      & FPR=0.1\%      & AUC                 & APCER              & BPCER              & ACER               \\ \hline
\multirow{3}{*}{1} & AENet \cite{zhang2020celeba} & 95.00          & 91.40          & 73.60          & 99.50               & 4.09               & 2.09               & 3.09               \\
                   & DPM \cite{zhang2022robust}             & \textbf{96.70} & 93.90          & 79.00          & -                   & \textbf{2.09}      & \textbf{1.57}      & \textbf{1.83}      \\
                   \rowcolor[HTML]{EFEFEF}
                   & Ours             & 95.72          & \textbf{94.01} & \textbf{90.39} & \textbf{99.75}      & 2.43               & 2.40               & 2.42               \\ \hline
\multirow{3}{*}{2} & AENet \cite{zhang2020celeba} & -              & -              & -              & 99.80±0.00         & 4.94±3.42          & 1.24±0.73          & 3.09±2.08          \\
                   & DPM \cite{zhang2022robust}              & -              & -              & -              & -                   & 1.02±0.29          & \textbf{0.72±0.62} & \textbf{0.73±0.64} \\
                   \rowcolor[HTML]{EFEFEF}
                   & Ours             & -              & -              & -              & \textbf{99.89±0.16} & \textbf{0.89±1.06} & 0.91±1.09          & 0.90±1.07          \\ \hline
\end{tabular}%
}
\end{table*}

\begin{table*}[]
\caption{Inference time comparison between two different methods.}
\label{fig:speed}
\centering
\setlength\tabcolsep{8pt}
\resizebox{2\columnwidth}{!}{%
\begin{tabular}{c|c|c|c|c|c|c}
\hline
Method  & Backbones       & Flops (G)     & Params (M)    & Speed (FPS) & Avg HTER (\%) & Avg AUC (\%) \\ \hline
FRT-PAD \cite{zhang2022effective} & ResNet-18 + GAT & 3.01          & 75.56         & 53          & 10.24         & \textbf{95.84}        \\
\rowcolor[HTML]{EFEFEF}
Ours    & ResNet-18       & \textbf{1.82} & \textbf{11.70} & \textbf{95} & \textbf{9.79}      & 95.83        \\ \hline
\end{tabular}%
}
\end{table*}

\subsection{Intra-Dataset Testing}

To validate the effectiveness of the proposed method, we conducted experiments on CelebA-Spoof \cite{zhang2020celeba} for intra-dataset testing. Subsequently, we compare the obtained results with other face anti-spoofing methods.

The proposed domain adversarial attack method is designed to facilitate the learning of domain-invariant features. However, during intra-dataset testing, where only one domain exists, the application of DAA is not feasible. Therefore, for intra-dataset testing, we exclusively employ a dual-teacher knowledge distillation method. This method integrates both face recognition and face attribute editing teacher models. The training process, guided by Eq. \eqref{eq:l-sum}, utilizes clean images as input for both the student and teachers.

Two protocols for intra-dataset testing, specifically designed to evaluate the performance of face anti-spoofing methods, are shown in Table \ref{tab:inta}. Additionally, we conducted a comparative comparison between the proposed method and other established methods, namely ACNet \cite{zhang2020celeba} and DPM \cite{zhang2022robust}. The comparison results with other methods are summarized in Table \ref{tab:intra-comp}. 
Specifically, the proposed method outperforms the state-of-the-art method across several evaluation metrics, substantiating the effectiveness of our approach.

\subsection{Comparisons of Different Architectures}


\begin{figure*}[!h]
  \centering
  \includegraphics[width=.85\textwidth]{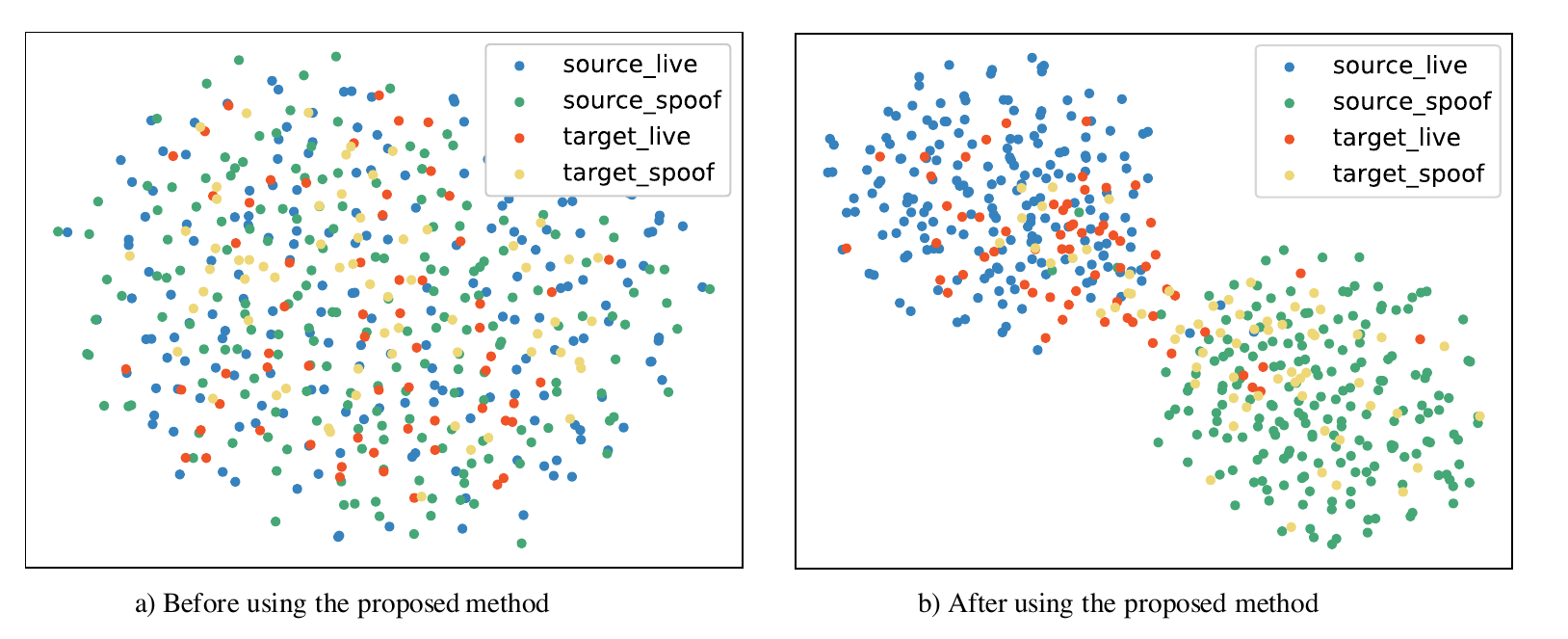}
  \vspace{-10pt}
  \caption{\textbf{The t-SNE visualizations of the extracted features under O\&M\&I to C.} a) Visualization results before using the DTDA method. b) Visualization results after using the DTDA method. Distinct colors are used to signify features originating from live images, spoof images, or features derived from different domains.
  }
  \label{fig:tsne}
\end{figure*}

Since the proposed method and \cite{zhang2022effective} all leverage other face-related tasks. We compare two different architectures of feature encoders to investigate the computational complexity of the proposed method. Comparison results of different network backbones, network parameters, FLOPs (floating point operations), and time cost are shown in Table \ref{fig:speed}. RTK-PAD and our method all adopt ResNet-18 as the backbone, but FRT-PAD uses an extra Graph Attention Network (GAT), which has increased the computational complexity and inference time. Thus, 1.19 G Flops and 63.86 M parameters can be saved by the proposed method. The inference time of each method is tested on a single GPU with 256 × 256 image resolution. And the proposed method can achieve 95 FPS, almost 1.8 times faster than \cite{zhang2022effective}. This also indicates that GAT is time-consuming. Specifically, the Avg HTER and AUC represent the average results of four testing tasks. The proposed method is more suitable for real-world applications than FRT-PAD, not only in inference time but also in accuracy. Other different network architectures can also leverage the proposed method for face knowledge distillation, and they may achieve better performance.

\begin{figure*}[]
  \centering
  \includegraphics[width=.90\textwidth]{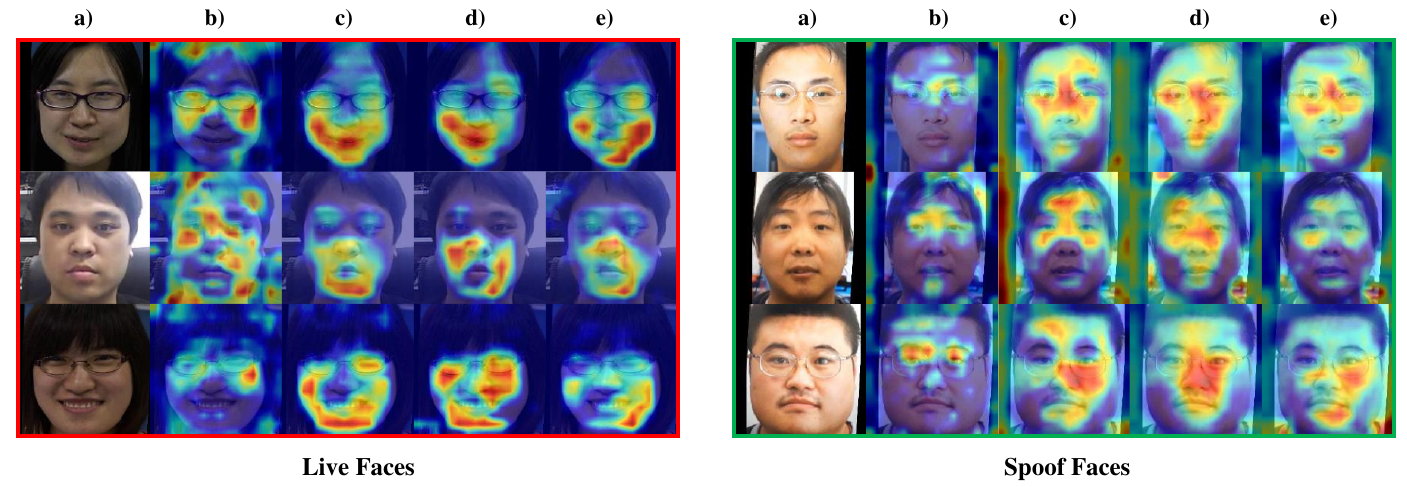}
  \caption{\textbf{Grad-CAM visualizations of the proposed method under O\&M\&I to C.} The images in the red box show the visualization of live faces, and the green box displays the visualization results of spoof faces.  a) Original image. b) Visualizations after using the ImageNet pre-train model. c) Visualization results after using the face attribute editing model. d) Visualization results after using the face recognition model. e) Visualization results after combining face attribute editing and recognition models.
  }
  \label{fig:cam}
\end{figure*}

\begin{figure*}[]
  \centering
  \includegraphics[width=.80\textwidth]{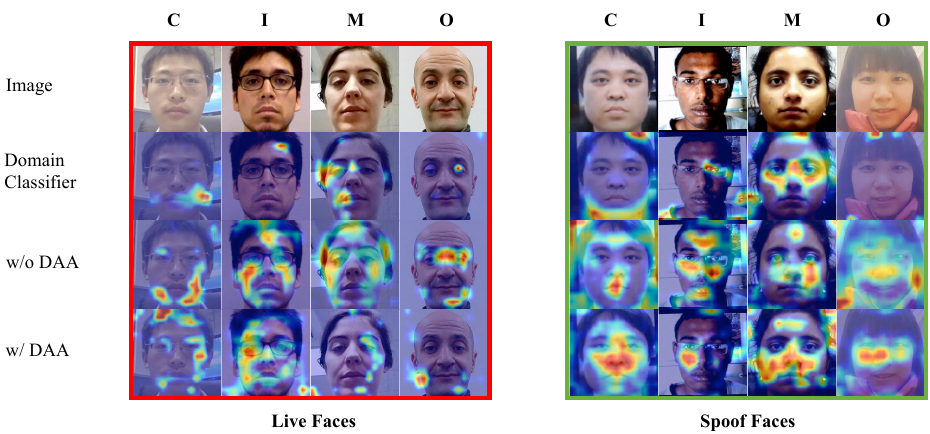}
  \caption{\textbf{Grad-CAM visualizations of the proposed DAA method on four datasets.} The four datasets are denoted as C, I, M, and O, respectively. The images in the red box show the visualization of live faces, and the green box displays the visualization results of spoof faces. The first row shows the original images, while the second row displays the activated regions of the domain classifier. The activated regions without using DAA and with using DAA are shown in the third and fourth rows, respectively.
  }
  \label{fig:cam2}
\end{figure*}

\subsection{Visualization}

\textbf{t-SNE Visualization}. To better illustrate our method, we analyze the feature space learned by our DTDA method. Specifically, we randomly select $300$ source samples and $100$ target samples from the four datasets and visualize the distribution of different features using t-SNE \cite{van2008visualizing}. As shown in Fig. \ref{fig:tsne}, before using the proposed method, all the features, whether from live or spoof images, are randomly distributed in the feature space.
However, by using both dual teacher and DAA, the distribution between live and spoof images becomes more dispersed, resulting in a better class boundary in the target domain. This improvement demonstrates the effectiveness of the proposed method.

\textbf{Grad-CAM Visualization}. To further investigate the effectiveness of the proposed dual teacher method and domain adversarial attack method, we utilize Grad-CAM \cite{selvaraju2017grad} to generate activation maps for both methods on the original images. 
The activation maps for the dual teacher module are shown in Fig. \ref{fig:cam}, which are generated using the O\&M\&I to C protocol.
The first row of the figure shows the original image, indicating that when using network weights pre-trained on ImageNet, some activated regions are in the background. However, after using either the face attribute editing network or the face recognition network, the activated regions mainly focus on the landmark areas in faces, regardless of whether the images are live or spoofed. Furthermore, when combining feature attribute editing and face recognition networks, the proposed method consistently focuses on the relevant region of interest, such as the mouth or eyes region of the face, to extract discriminative cues instead of domain-specific backgrounds. This approach is more likely to generalize well to unseen domains.

In addition, we also visualize the effectiveness of the domain adversarial attack (DAA) method. The activation results of the domain classifier are shown in the second row of Fig. \ref{fig:cam2}, where the activated regions contain domain-specific features.
The first row of the figure shows the original image. Without using DAA, there may be some overlapping activated regions with the domain classifier regions in the third row. However, after using DAA, those domain-related activations become inactive, indicating that DAA has forced the network to focus on domain-specific regions rather than domain-independent regions. By using DAA, the network can learn domain-invariant features, which can improve the generalization of the model.

\section{Conclusion}

In this paper, we propose a framework of dual teacher knowledge distillation with domain alignment for face anti-spoofing to improve the generalization performance of the model.
We first propose a domain adversarial attack method that aims to extract domain-invariant features.
To further acquire more face priors from other face tasks, we propose a dual teacher model for knowledge distillation.
The effectiveness of the proposed method is verified by extensive ablation studies and comparison with SOTA methods on public datasets with various protocols.

\balance

{\small
\bibliographystyle{IEEEtran}
\bibliography{sample-base}

\begin{thebibliography}{10}
\providecommand{\url}[1]{#1}
\csname url@samestyle\endcsname
\providecommand{\newblock}{\relax}
\providecommand{\bibinfo}[2]{#2}
\providecommand{\BIBentrySTDinterwordspacing}{\spaceskip=0pt\relax}
\providecommand{\BIBentryALTinterwordstretchfactor}{4}
\providecommand{\BIBentryALTinterwordspacing}{\spaceskip=\fontdimen2\font plus
\BIBentryALTinterwordstretchfactor\fontdimen3\font minus \fontdimen4\font\relax}
\providecommand{\BIBforeignlanguage}[2]{{%
\expandafter\ifx\csname l@#1\endcsname\relax
\typeout{** WARNING: IEEEtran.bst: No hyphenation pattern has been}%
\typeout{** loaded for the language `#1'. Using the pattern for}%
\typeout{** the default language instead.}%
\else
\language=\csname l@#1\endcsname
\fi
#2}}
\providecommand{\BIBdecl}{\relax}
\BIBdecl

\bibitem{deng2019arcface}
J.~Deng, J.~Guo, N.~Xue, and S.~Zafeiriou, ``Arcface: Additive angular margin loss for deep face recognition,'' in \emph{Proceedings of the IEEE Conference on Computer Vision and Pattern Recognition}, 2019, pp. 4690--4699.

\bibitem{de2013can}
T.~de~Freitas~Pereira, A.~Anjos, J.~M. De~Martino, and S.~Marcel, ``Can face anti-spoofing countermeasures work in a real world scenario?'' in \emph{Proceedings of the International Conference on Biometrics}.\hskip 1em plus 0.5em minus 0.4em\relax IEEE, 2013, pp. 1--8.

\bibitem{yang2013face}
J.~Yang, Z.~Lei, S.~Liao, and S.~Z. Li, ``Face liveness detection with component dependent descriptor,'' in \emph{Proceedings of the International Conference on Biometrics}.\hskip 1em plus 0.5em minus 0.4em\relax IEEE, 2013, pp. 1--6.

\bibitem{patel2016secure}
K.~Patel, H.~Han, and A.~K. Jain, ``Secure face unlock: Spoof detection on smartphones,'' \emph{IEEE Transactions on Information Forensics and Security}, vol.~11, no.~10, pp. 2268--2283, 2016.

\bibitem{boulkenafet2016face}
Z.~Boulkenafet, J.~Komulainen, and A.~Hadid, ``Face antispoofing using speeded-up robust features and fisher vector encoding,'' \emph{IEEE Signal Processing Letters}, vol.~24, no.~2, pp. 141--145, 2016.

\bibitem{wang2020cross}
G.~Wang, H.~Han, S.~Shan, and X.~Chen, ``Cross-domain face presentation attack detection via multi-domain disentangled representation learning,'' in \emph{Proceedings of the IEEE Conference on Computer Vision and Pattern Recognition}, 2020, pp. 6678--6687.

\bibitem{yu2020searching}
Z.~Yu, C.~Zhao, Z.~Wang, Y.~Qin, Z.~Su, X.~Li, F.~Zhou, and G.~Zhao, ``Searching central difference convolutional networks for face anti-spoofing,'' in \emph{Proceedings of the IEEE Conference on Computer Vision and Pattern Recognition}, 2020, pp. 5295--5305.

\bibitem{maatta2011face}
J.~M{\"a}{\"a}tt{\"a}, A.~Hadid, and M.~Pietik{\"a}inen, ``Face spoofing detection from single images using micro-texture analysis,'' in \emph{Proceedings of the International Joint Conference on Biometrics}.\hskip 1em plus 0.5em minus 0.4em\relax IEEE, 2011, pp. 1--7.

\bibitem{wen2015face}
D.~Wen, H.~Han, and A.~K. Jain, ``Face spoof detection with image distortion analysis,'' \emph{IEEE Transactions on Information Forensics and Security}, vol.~10, no.~4, pp. 746--761, 2015.

\bibitem{Liu_2018_CVPR}
Y.~Liu, A.~Jourabloo, and X.~Liu, ``Learning deep models for face anti-spoofing: Binary or auxiliary supervision,'' in \emph{Proceedings of the IEEE Conference on Computer Vision and Pattern Recognition}, June 2018.

\bibitem{Wang_2022_CVPR}
Z.~Wang, Z.~Wang, Z.~Yu, W.~Deng, J.~Li, T.~Gao, and Z.~Wang, ``Domain generalization via shuffled style assembly for face anti-spoofing,'' in \emph{Proceedings of the IEEE Conference on Computer Vision and Pattern Recognition}, June 2022, pp. 4123--4133.

\bibitem{shao2019multi}
R.~Shao, X.~Lan, J.~Li, and P.~C. Yuen, ``Multi-adversarial discriminative deep domain generalization for face presentation attack detection,'' in \emph{Proceedings of the IEEE Conference on Computer Vision and Pattern Recognition}, 2019, pp. 10\,023--10\,031.

\bibitem{qin2020learning}
Y.~Qin, C.~Zhao, X.~Zhu, Z.~Wang, Z.~Yu, T.~Fu, F.~Zhou, J.~Shi, and Z.~Lei, ``Learning meta model for zero-and few-shot face anti-spoofing,'' in \emph{Proceedings of the AAAI Conference on Artificial Intelligence}, vol.~34, no.~07, 2020, pp. 11\,916--11\,923.

\bibitem{shao2020regularized}
R.~Shao, X.~Lan, and P.~C. Yuen, ``Regularized fine-grained meta face anti-spoofing,'' in \emph{Proceedings of the AAAI Conference on Artificial Intelligence}, vol.~34, no.~07, 2020, pp. 11\,974--11\,981.

\bibitem{guo2022multi}
X.~Guo, Y.~Liu, A.~Jain, and X.~Liu, ``Multi-domain learning for updating face anti-spoofing models,'' in \emph{Proceedings of the European Conference on Computer Vision}.\hskip 1em plus 0.5em minus 0.4em\relax Springer, 2022, pp. 230--249.

\bibitem{sun2023rethinking}
Y.~Sun, Y.~Liu, X.~Liu, Y.~Li, and W.-S. Chu, ``Rethinking domain generalization for face anti-spoofing: Separability and alignment,'' in \emph{Proceedings of the IEEE Conference on Computer Vision and Pattern Recognition}, 2023, pp. 24\,563--24\,574.

\bibitem{liu2023towards}
Y.~Liu, Y.~Chen, M.~Gou, C.-T. Huang, Y.~Wang, W.~Dai, and H.~Xiong, ``Towards unsupervised domain generalization for face anti-spoofing,'' in \emph{IEEE International Conference on Computer Vision}, 2023, pp. 20\,654--20\,664.

\bibitem{yan2022domain}
W.~Yan, Y.~Zeng, and H.~Hu, ``Domain adversarial disentanglement network with cross-domain synthesis for generalized face anti-spoofing,'' \emph{IEEE Transactions on Circuits and Systems for Video Technology}, vol.~32, no.~10, pp. 7033--7046, 2022.

\bibitem{Jia_2020_CVPR}
Y.~Jia, J.~Zhang, S.~Shan, and X.~Chen, ``Single-side domain generalization for face anti-spoofing,'' in \emph{Proceedings of the IEEE Conference on Computer Vision and Pattern Recognition}, June 2020.

\bibitem{ganin2015unsupervised}
Y.~Ganin and V.~Lempitsky, ``Unsupervised domain adaptation by backpropagation,'' in \emph{Proceedings of the International Conference on Machine Learning}.\hskip 1em plus 0.5em minus 0.4em\relax PMLR, 2015, pp. 1180--1189.

\bibitem{zhang2023free}
Y.~Zhang, X.~Wang, J.~Liang, Z.~Zhang, L.~Wang, R.~Jin, and T.~Tan, ``Free lunch for domain adversarial training: Environment label smoothing,'' \emph{arXiv preprint arXiv:2302.00194}, 2023.

\bibitem{hinton2015distilling}
G.~Hinton, O.~Vinyals, and J.~Dean, ``Distilling the knowledge in a neural network,'' \emph{arXiv preprint arXiv:1503.02531}, 2015.

\bibitem{liu2022fingerprint}
F.~Liu, Z.~Kong, H.~Liu, W.~Zhang, and L.~Shen, ``Fingerprint presentation attack detection by channel-wise feature denoising,'' \emph{IEEE Transactions on Information Forensics and Security}, vol.~17, pp. 2963--2976, 2022.

\bibitem{Liu_2018_ECCV}
S.-Q. Liu, X.~Lan, and P.~C. Yuen, ``Remote photoplethysmography correspondence feature for 3d mask face presentation attack detection,'' in \emph{Proceedings of the European Conference on Computer Vision}, September 2018.

\bibitem{liu20163d}
S.~Liu, P.~C. Yuen, S.~Zhang, and G.~Zhao, ``3d mask face anti-spoofing with remote photoplethysmography,'' in \emph{Proceedings of the European Conference on Computer Vision}.\hskip 1em plus 0.5em minus 0.4em\relax Springer, 2016, pp. 85--100.

\bibitem{yu2020face}
Z.~Yu, X.~Li, X.~Niu, J.~Shi, and G.~Zhao, ``Face anti-spoofing with human material perception,'' in \emph{Proceedings of the European Conference on Computer Vision}.\hskip 1em plus 0.5em minus 0.4em\relax Springer, 2020, pp. 557--575.

\bibitem{yang2019face}
X.~Yang, W.~Luo, L.~Bao, Y.~Gao, D.~Gong, S.~Zheng, Z.~Li, and W.~Liu, ``Face anti-spoofing: Model matters, so does data,'' in \emph{Proceedings of the IEEE Conference on Computer Vision and Pattern Recognition}, 2019, pp. 3507--3516.

\bibitem{zhang2022effective}
W.~Zhang, H.~Liu, F.~Liu, R.~Ramachandra, and C.~Busch, ``Effective presentation attack detection driven by face related task,'' in \emph{Proceedings of the European Conference on Computer Vision}.\hskip 1em plus 0.5em minus 0.4em\relax Springer, 2022, pp. 408--423.

\bibitem{wu2021dual}
H.~Wu, D.~Zeng, Y.~Hu, H.~Shi, and T.~Mei, ``Dual spoof disentanglement generation for face anti-spoofing with depth uncertainty learning,'' \emph{IEEE Transactions on Circuits and Systems for Video Technology}, vol.~32, no.~7, pp. 4626--4638, 2021.

\bibitem{arashloo2020unseen}
S.~R. Arashloo, ``Unseen face presentation attack detection using sparse multiple kernel fisher null-space,'' \emph{IEEE Transactions on Circuits and Systems for Video Technology}, vol.~31, no.~10, pp. 4084--4095, 2020.

\bibitem{yu2020fas}
Z.~Yu, J.~Wan, Y.~Qin, X.~Li, S.~Z. Li, and G.~Zhao, ``Nas-fas: Static-dynamic central difference network search for face anti-spoofing,'' \emph{IEEE Transactions on Pattern Analysis and Machine Intelligence}, vol.~43, no.~9, pp. 3005--3023, 2020.

\bibitem{liu2020disentangling}
Y.~Liu, J.~Stehouwer, and X.~Liu, ``On disentangling spoof trace for generic face anti-spoofing,'' in \emph{Proceedings of the European Conference on Computer Vision}.\hskip 1em plus 0.5em minus 0.4em\relax Springer, 2020, pp. 406--422.

\bibitem{zhang2020face}
K.-Y. Zhang, T.~Yao, J.~Zhang, Y.~Tai, S.~Ding, J.~Li, F.~Huang, H.~Song, and L.~Ma, ``Face anti-spoofing via disentangled representation learning,'' in \emph{Proceedings of the European Conference on Computer Vision}.\hskip 1em plus 0.5em minus 0.4em\relax Springer, 2020, pp. 641--657.

\bibitem{chen2021generalizable}
Z.~Chen, T.~Yao, K.~Sheng, S.~Ding, Y.~Tai, J.~Li, F.~Huang, and X.~Jin, ``Generalizable representation learning for mixture domain face anti-spoofing,'' in \emph{Proceedings of the AAAI Conference on Artificial Intelligence}, vol.~35, no.~2, 2021, pp. 1132--1139.

\bibitem{qin2021meta}
Y.~Qin, Z.~Yu, L.~Yan, Z.~Wang, C.~Zhao, and Z.~Lei, ``Meta-teacher for face anti-spoofing,'' \emph{IEEE Transactions on Pattern Analysis and Machine Intelligence}, vol.~44, no.~10, pp. 6311--6326, 2021.

\bibitem{wang2021self}
J.~Wang, J.~Zhang, Y.~Bian, Y.~Cai, C.~Wang, and S.~Pu, ``Self-domain adaptation for face anti-spoofing,'' in \emph{Proceedings of the AAAI Conference on Artificial Intelligence}, vol.~35, no.~4, 2021, pp. 2746--2754.

\bibitem{10051654}
Z.~Kong, W.~Zhang, F.~Liu, W.~Luo, H.~Liu, L.~Shen, and R.~Ramachandra, ``Taming self-supervised learning for presentation attack detection: De-folding and de-mixing,'' \emph{IEEE Transactions on Neural Networks and Learning Systems}, pp. 1--12, 2023.

\bibitem{wang2022patchnet}
C.-Y. Wang, Y.-D. Lu, S.-T. Yang, and S.-H. Lai, ``Patchnet: A simple face anti-spoofing framework via fine-grained patch recognition,'' in \emph{Proceedings of the IEEE Conference on Computer Vision and Pattern Recognition}, 2022, pp. 20\,281--20\,290.

\bibitem{wang2023consistency}
Z.~Wang, Z.~Yu, X.~Wang, Y.~Qin, J.~Li, C.~Zhao, X.~Liu, and Z.~Lei, ``Consistency regularization for deep face anti-spoofing,'' \emph{IEEE Transactions on Information Forensics and Security}, 2023.

\bibitem{szegedy2013intriguing}
C.~Szegedy, W.~Zaremba, I.~Sutskever, J.~Bruna, D.~Erhan, I.~Goodfellow, and R.~Fergus, ``Intriguing properties of neural networks,'' \emph{arXiv preprint arXiv:1312.6199}, 2013.

\bibitem{madry2017towards}
A.~Madry, A.~Makelov, L.~Schmidt, D.~Tsipras, and A.~Vladu, ``Towards deep learning models resistant to adversarial attacks,'' \emph{arXiv preprint arXiv:1706.06083}, 2017.

\bibitem{goodfellow2014explaining}
I.~J. Goodfellow, J.~Shlens, and C.~Szegedy, ``Explaining and harnessing adversarial examples,'' \emph{arXiv preprint arXiv:1412.6572}, 2014.

\bibitem{carlini2017towards}
N.~Carlini and D.~Wagner, ``Towards evaluating the robustness of neural networks,'' in \emph{Proceedings of the IEEE Symposium on Security and Privacy}.\hskip 1em plus 0.5em minus 0.4em\relax IEEE, 2017, pp. 39--57.

\bibitem{dong2018boosting}
Y.~Dong, F.~Liao, T.~Pang, H.~Su, J.~Zhu, X.~Hu, and J.~Li, ``Boosting adversarial attacks with momentum,'' in \emph{Proceedings of the IEEE Conference on Computer Vision and Pattern Recognition}, 2018, pp. 9185--9193.

\bibitem{zagoruyko2016paying}
S.~Zagoruyko and N.~Komodakis, ``Paying more attention to attention: Improving the performance of convolutional neural networks via attention transfer,'' \emph{arXiv preprint arXiv:1612.03928}, 2016.

\bibitem{park2019relational}
W.~Park, D.~Kim, Y.~Lu, and M.~Cho, ``Relational knowledge distillation,'' in \emph{Proceedings of the IEEE Conference on Computer Vision and Pattern Recognition}, 2019, pp. 3967--3976.

\bibitem{tung2019similarity}
F.~Tung and G.~Mori, ``Similarity-preserving knowledge distillation,'' in \emph{Proceedings of the International Conference on Computer Vision}, 2019, pp. 1365--1374.

\bibitem{romero2014fitnets}
A.~Romero, N.~Ballas, S.~E. Kahou, A.~Chassang, C.~Gatta, and Y.~Bengio, ``Fitnets: Hints for thin deep nets,'' \emph{arXiv preprint arXiv:1412.6550}, 2014.

\bibitem{tian2019contrastive}
Y.~Tian, D.~Krishnan, and P.~Isola, ``Contrastive representation distillation,'' \emph{arXiv preprint arXiv:1910.10699}, 2019.

\bibitem{zhao2023learn}
Q.~Zhao, S.~Lyu, L.~Chen, B.~Liu, T.-B. Xu, G.~Cheng, and W.~Feng, ``Learn by oneself: Exploiting weight-sharing potential in knowledge distillation guided ensemble network,'' \emph{IEEE Transactions on Circuits and Systems for Video Technology}, 2023.

\bibitem{zhang2023unsupervised}
X.~Zhang, X.~Wang, and P.~Cheng, ``Unsupervised hashing retrieval via efficient correlation distillation,'' \emph{IEEE Transactions on Circuits and Systems for Video Technology}, 2023.

\bibitem{heo2019comprehensive}
B.~Heo, J.~Kim, S.~Yun, H.~Park, N.~Kwak, and J.~Y. Choi, ``A comprehensive overhaul of feature distillation,'' in \emph{Proceedings of the International Conference on Computer Vision}, 2019, pp. 1921--1930.

\bibitem{li2020face}
H.~Li, S.~Wang, P.~He, and A.~Rocha, ``Face anti-spoofing with deep neural network distillation,'' \emph{IEEE Journal of Selected Topics in Signal Processing}, vol.~14, no.~5, pp. 933--946, 2020.

\bibitem{li2022one}
Z.~Li, R.~Cai, H.~Li, K.-Y. Lam, Y.~Hu, and A.~C. Kot, ``One-class knowledge distillation for face presentation attack detection,'' \emph{IEEE Transactions on Information Forensics and Security}, vol.~17, pp. 2137--2150, 2022.

\bibitem{liu2022cross}
Y.~Liu, J.~Cao, B.~Li, W.~Hu, J.~Ding, and L.~Li, ``Cross-architecture knowledge distillation,'' in \emph{Proceedings of the Asian Conference on Computer Vision}, 2022, pp. 3396--3411.

\bibitem{dosovitskiy2020image}
A.~Dosovitskiy, L.~Beyer, A.~Kolesnikov, D.~Weissenborn, X.~Zhai, T.~Unterthiner, M.~Dehghani, M.~Minderer, G.~Heigold, S.~Gelly \emph{et~al.}, ``An image is worth 16x16 words: Transformers for image recognition at scale,'' \emph{arXiv preprint arXiv:2010.11929}, 2020.

\bibitem{zhang2021survey}
Y.~Zhang and Q.~Yang, ``A survey on multi-task learning,'' \emph{IEEE Transactions on Knowledge and Data Engineering}, vol.~34, no.~12, pp. 5586--5609, 2021.

\bibitem{boulkenafet2017oulu}
Z.~Boulkenafet, J.~Komulainen, L.~Li, X.~Feng, and A.~Hadid, ``Oulu-npu: A mobile face presentation attack database with real-world variations,'' in \emph{Proceedings of the IEEE International Conference on Automatic Face \& Gesture Recognition}.\hskip 1em plus 0.5em minus 0.4em\relax IEEE, 2017, pp. 612--618.

\bibitem{zhang2012face}
Z.~Zhang, J.~Yan, S.~Liu, Z.~Lei, D.~Yi, and S.~Z. Li, ``A face antispoofing database with diverse attacks,'' in \emph{Proceedings of the International Conference on Biometrics}.\hskip 1em plus 0.5em minus 0.4em\relax IEEE, 2012, pp. 26--31.

\bibitem{chingovska2012effectiveness}
I.~Chingovska, A.~Anjos, and S.~Marcel, ``On the effectiveness of local binary patterns in face anti-spoofing,'' in \emph{Proceedings of the International Conference of Biometrics Special Interest Group}.\hskip 1em plus 0.5em minus 0.4em\relax IEEE, 2012, pp. 1--7.

\bibitem{zhang2020celeba}
Y.~Zhang, Z.~Yin, Y.~Li, G.~Yin, J.~Yan, J.~Shao, and Z.~Liu, ``Celeba-spoof: Large-scale face anti-spoofing dataset with rich annotations,'' in \emph{Proceedings of the European Conference on Computer Vision}, 2020, pp. 70--85.

\bibitem{zhang2016joint}
K.~Zhang, Z.~Zhang, Z.~Li, and Y.~Qiao, ``Joint face detection and alignment using multitask cascaded convolutional networks,'' \emph{IEEE Signal Processing Letters}, vol.~23, no.~10, pp. 1499--1503, 2016.

\bibitem{he2016deep}
K.~He, X.~Zhang, S.~Ren, and J.~Sun, ``Deep residual learning for image recognition,'' in \emph{Proceedings of the IEEE Conference on Computer Vision and Pattern Recognition}, 2016, pp. 770--778.

\bibitem{choi2018stargan}
Y.~Choi, M.~Choi, M.~Kim, J.-W. Ha, S.~Kim, and J.~Choo, ``Stargan: Unified generative adversarial networks for multi-domain image-to-image translation,'' in \emph{Proceedings of the IEEE Conference on Computer Vision and Pattern Recognition}, 2018, pp. 8789--8797.

\bibitem{liu2021dual}
S.~Liu, K.-Y. Zhang, T.~Yao, K.~Sheng, S.~Ding, Y.~Tai, J.~Li, Y.~Xie, and L.~Ma, ``Dual reweighting domain generalization for face presentation attack detection,'' \emph{arXiv preprint arXiv:2106.16128}, 2021.

\bibitem{wang2021vlad}
J.~Wang, Z.~Zhao, W.~Jin, X.~Duan, Z.~Lei, B.~Huai, Y.~Wu, and X.~He, ``Vlad-vsa: cross-domain face presentation attack detection with vocabulary separation and adaptation,'' in \emph{Proceedings of the 29th ACM International Conference on Multimedia}, 2021, pp. 1497--1506.

\bibitem{liu2021adaptive}
S.~Liu, K.-Y. Zhang, T.~Yao, M.~Bi, S.~Ding, J.~Li, F.~Huang, and L.~Ma, ``Adaptive normalized representation learning for generalizable face anti-spoofing,'' in \emph{Proceedings of the 29th ACM International Conference on Multimedia}, 2021, pp. 1469--1477.

\bibitem{zhou2022generative}
Q.~Zhou, K.-Y. Zhang, T.~Yao, R.~Yi, K.~Sheng, S.~Ding, and L.~Ma, ``Generative domain adaptation for face anti-spoofing,'' in \emph{Proceedings of the European Conference on Computer Vision}.\hskip 1em plus 0.5em minus 0.4em\relax Springer, 2022, pp. 335--356.

\bibitem{zhang2022robust}
Y.~Zhang, Y.~Wu, Z.~Yin, J.~Shao, and Z.~Liu, ``Robust face anti-spoofing with dual probabilistic modeling,'' \emph{arXiv preprint arXiv:2204.12685}, 2022.

\bibitem{van2008visualizing}
L.~Van~der Maaten and G.~Hinton, ``Visualizing data using t-sne.'' \emph{Journal of Machine Learning Research}, vol.~9, no.~11, 2008.

\bibitem{selvaraju2017grad}
R.~R. Selvaraju, M.~Cogswell, A.~Das, R.~Vedantam, D.~Parikh, and D.~Batra, ``Grad-cam: Visual explanations from deep networks via gradient-based localization,'' in \emph{Proceedings of the IEEE Conference on Computer Vision and Pattern Recognition}, 2017, pp. 618--626.

\end{thebibliography}
}

\end{document}